\def\year{2022}\relax
\documentclass[letterpaper]{article} % DO NOT CHANGE THIS
\usepackage{aaai22}  % DO NOT CHANGE THIS
\usepackage{times}  % DO NOT CHANGE THIS
\usepackage{helvet}  % DO NOT CHANGE THIS
\usepackage{courier}  % DO NOT CHANGE THIS
\usepackage[hyphens]{url}  % DO NOT CHANGE THIS
\usepackage{graphicx} % DO NOT CHANGE THIS
\usepackage{natbib}  % DO NOT CHANGE THIS AND DO NOT ADD ANY OPTIONS TO IT
\usepackage{caption} % DO NOT CHANGE THIS AND DO NOT ADD ANY OPTIONS TO IT
\DeclareCaptionStyle{ruled}{labelfont=normalfont,labelsep=colon,strut=off} % DO NOT CHANGE THIS
\usepackage{multirow}
\usepackage{algorithmicx,algorithm}
\usepackage[noend]{algpseudocode}
\usepackage{multirow}
\usepackage{amsmath}
\usepackage{color}

\usepackage{amssymb}

\usepackage{newfloat}
\usepackage{listings}
\floatstyle{ruled}
\newfloat{listing}{tb}{lst}{}
\floatname{listing}{Listing}

\title{Student Helping Teacher: Teacher Evolution via Self-Knowledge Distillation}

\makeatletter% <-------------------
\renewcommand{\copyright@on}{}
\makeatother

\author {
	Zheng Li,\textsuperscript{\rm 1,4}\thanks{This work is done when Zheng Li is a research intern at Megvii}
	Xiang Li, \textsuperscript{\rm 2}
	Lingfeng Yang, \textsuperscript{\rm 2,4}
	Jian Yang, \textsuperscript{\rm 2} 
	Zhigeng Pan \textsuperscript{\rm 3}\thanks{Corresponding author}
	\\
}
\affiliations {
	\textsuperscript{\rm 1} Hangzhou Normal University \\
	\textsuperscript{\rm 2} Nanjing Unversity of Science and Technology \\
	\textsuperscript{\rm 3} Nanjing Unversity of Information Science and Technology \\
	\textsuperscript{\rm 4} MEGVII Technology \\
	lizheng1@stu.hznu.edu.cn
}
\usepackage{bibentry}
\begin{document}

\maketitle

\begin{abstract}
	
Knowledge distillation usually transfers the knowledge from a pre-trained cumbersome teacher network to a compact student network, which follows the classical \textit{teacher-teaching-student} paradigm. Based on this paradigm, previous methods mostly focus on how to efficiently train a better student network for deployment. Different from the existing practices, in this paper, we propose a novel \textit{student-helping-teacher} formula, Teacher Evolution via Self-Knowledge Distillation (TESKD), where the target teacher (for deployment) is learned with the help of multiple hierarchical students by sharing the structural backbone. The diverse feedback from multiple students allows the teacher to improve itself through the shared feature representations.
%reflect on himself and get a significant improvement.
The effectiveness of our proposed framework is demonstrated by extensive experiments with various network settings on two standard benchmarks including CIFAR-100 and ImageNet. Notably, when trained together with our proposed method, ResNet-18 achieves 79.15\% and 71.14\% accuracy on CIFAR-100 and ImageNet, outperforming the baseline results by 4.74\% and 1.43\%, respectively. The code is available at: \url{https://github.com/zhengli427/TESKD}.
	
%\rewrite{TODO: simply explain why it works conceptually.}
%with better generalization ability and use it for deployment.	% The well-trained student performs well in both accuracy and efficiency.	% Another student-teaching-student distillation paradigm has also been proposed to train a pool of students simultaneously in a peer-teaching manner and use the well-trained student for deployment. 
	
% Rather than compressing the teacher to obtain a compact student, in contrast, in this paper, we propose a novel \textit{student-helping-teacher} paradigm, \lz{paper title}, which trains the teacher with the help of multiple students. % and use the well-trained teacher for final deployment. 
	
\end{abstract}

\section{Introduction}

% BYOT 存在的问题就是auxiliary classifier里面，
% 再画一张图去对比我们的方法和BYOT还有KD的结构方式。

\begin{figure}[t]
	\centering
	\includegraphics[width=1\linewidth]{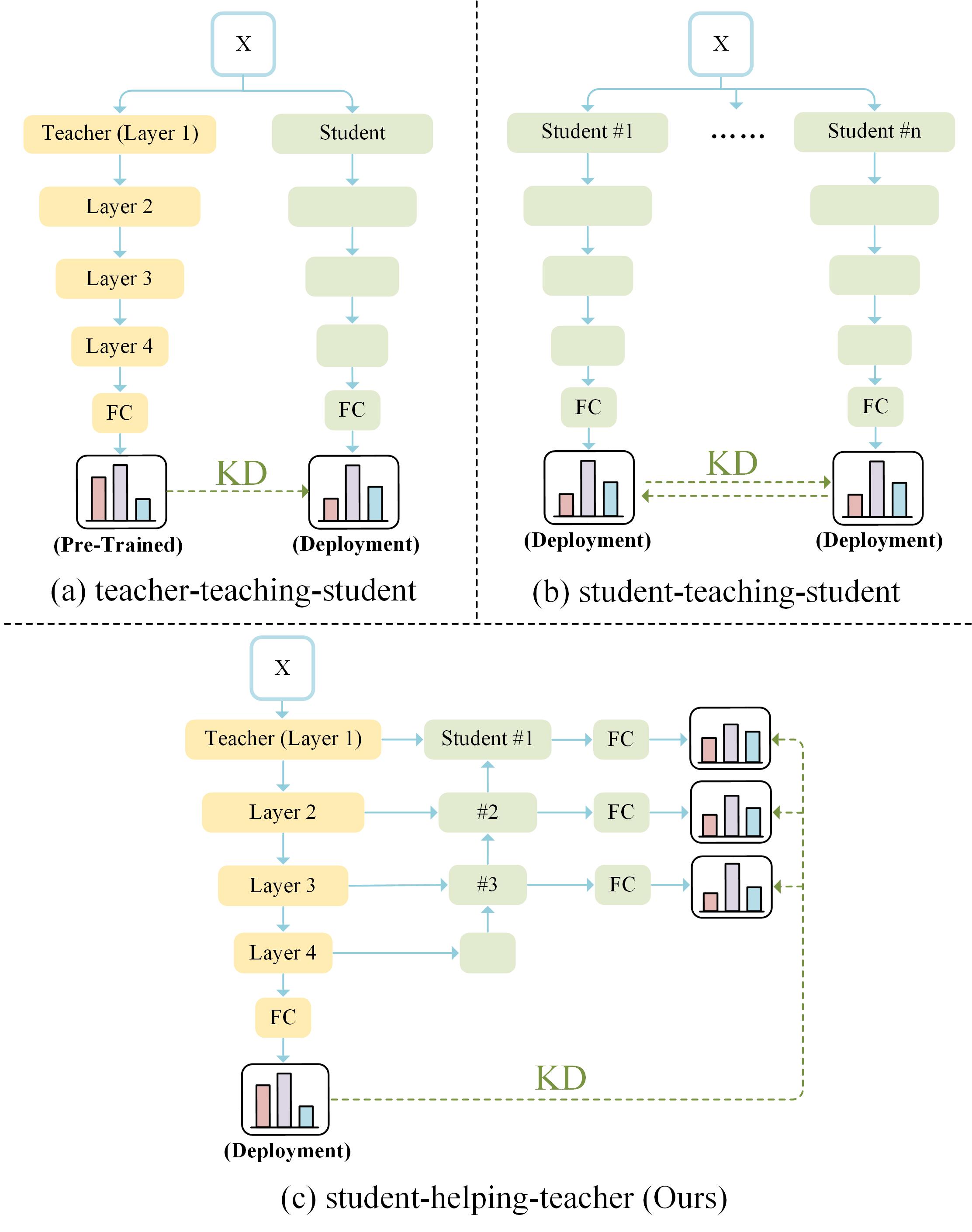}
	\caption{
		Comparison of three distillation paradigms. The blue line is the forward path and the dashed green line is the soft label distillation. The arrow points to the distilled target. As opposed to previous approaches that follow the classical \textit{teacher-teaching-student} or \textit{student-teaching-student} paradigm, we propose a novel \textit{student-helping-teacher} distillation paradigm. When the target teacher distills knowledge to the students, it can also be significantly improved based on the students' diverse feedback via shared intermediate representations.
		We omit the supervised learning loss and feature distillation loss for simplicity.
	}
	\label{fig:comparison}
\end{figure}

Deep convolutional networks~(CNNs) have achieved remarkable success in various computer vision applications, including image classification~\cite{he2016deep,huang2017densely,xie2017aggregated,li2019selective}, object detection~\cite{girshick2015fast,ren2015faster,tian2019fcos,li2020generalized,li2021generalized}, semantic segmentation~\cite{long2015fully,ronneberger2015u} and pose estimation~\cite{newell2016stacked,chu2017multi}. With the growing number of model parameters, a large amount of computational resource is required to achieve state-of-the-art accuracy. However, networks with millions of parameters are hard to be deployed to platforms with limited computing resources. To address this issue, a variety of network compression approaches such as quantization~\cite{chen2015compressing,wu2016quantized}, network pruning~\cite{molchanov2016pruning,li2016pruning} and knowledge distillation~\cite{hinton2015distilling}, have been exploited to obtain a small network that can work as well as the large network while effectively reducing the computational costs and memory consumption.

% \rewrite{Check here.} % 句子避免太长，以2-3行为主。该拆开来讲的拆开来说。参考kaiming的一作论文。并且表达上的上下文连贯性需要注意。
Knowledge distillation, as one of the main network compression techniques, becomes increasingly popular recently. It usually transfers the knowledge of a cumbersome pre-trained teacher network in the form of soft predictions~\cite{hinton2015distilling} or intermediate representations~\cite{romero2014fitnets,zagoruyko2016paying,yim2017gift}, aiming at improving the generalization ability of a compact student network. Such a learning process can be typically viewed as a \textit{teacher-teaching-student} paradigm, where the fixed strong teacher teaches weak student and the well-trained student is used for final deployment, as shown in Fig.~\ref{fig:comparison}(a). 
Instead of knowledge transfer between a static teacher and a compact student, another distillation paradigm \textit{student-teaching-student} has also been proposed in DML~\cite{zhang2018deep}. In this paradigm, the pre-trained teacher network no longer exists, and all the individual networks are treated as students. Two or more students are trained simultaneously in a cooperative peer-teaching manner and gain extra knowledge from each other. Such a one-stage learning process significantly improve the training efficiency. In test, the best student is selected for final deployment. The whole mutual learning procedure is shown in Fig.~\ref{fig:comparison}(b). 

Different from the existing paradigms, we propose a novel \textit{student-helping-teacher} formula, \textbf{T}eacher \textbf{E}volution via \textbf{S}elf-\textbf{K}nowledge \textbf{D}istillation~(TESKD), which introduces multiple hierarchical classifiers~(students) to facilitate the learning of the backbone network~(teacher) for deployment, as illustrated in Fig.~\ref{fig:comparison}(c).
Specifically, we design a backbone teacher network with target complexity for deployment, and construct multiple student sub-networks in a FPN-like \cite{lin2017feature} way by sharing various stages of teacher backbone features. During training, once the teacher provides high-quality soft labels to guide the hierarchical students, it also offers the opportunity for the teacher to make meaningful improvements based on students' diverse feedback via the shared intermediate representations. In order to obtain the effective feedback, we propose the Mixed Fusion Module~(MFM) to build hierarchical student sub-networks with a top-down architecture. Specifically, MFM consists of both addition and concatenation operators, which diversely and sufficiently bridge the information flow between the teacher and multiple students.

The overall contributions of this paper are summarized as follows:
\begin{itemize}
	\item To our best knowledge, we are the first to propose the student-helping-teacher paradigm in knowledge distillation area. The target deployed teacher can make improvements by learning from the feedback of multiple hierarchical students, through their shared backbone features in our proposed one-stage self-distillation framework.
	%learn from students' feedback and benefit from it. The well-trained teacher will be used for final deployment.
	%	\item \rewrite{please write about MFM}
	\item To further improve the effectiveness of the feedback, we propose the Mixed Fusion Module (MFM) to build multiple hierarchical student sub-networks with a top-down architecture.
	%We propose a novel self-distillation network that follows the learning process of student-helping-teacher paradigm.
	%	\item The mixed fusion module is designed to build stronger students.
	\item Extensive experiments on CIFAR-100 and ImageNet-2012 datasets 
	%with a variety of settings 
	based on four popular network architectures demonstrate the effectiveness of our proposed self-distillation framework.
\end{itemize}

% The higher accuracy of the auxiliary classifiers, the stronger representation ability of the backbone network. 

% For such branch-based self-distillation method, BYOT neglect the fact that the higher accuracy of the auxiliary classifier, the stronger representation ability of the backbone network. The shared backbone network will learn more generalized features during the learning process of auxiliary classifiers, since all branch networks shares the same convolutional blocks of the backbone network. This will bring a significant improvement to the backbone network and outperforms the baseline network. The student networks are actually helping the teacher network. However, limited by the representation ability of the input features, the accuracy of the auxiliary label is not high enough, which limits the further improvement of the distillation performance.

\section{Related Work}

\textbf{Knowledge Distillation} Knowledge distillation aims at effectively learning a compact and comparable student model by transferring the knowledge of a pretrained cumbersome teacher model, which follows the classical teacher-teaching-student distillation paradigm. Hinton \textit{et al.}~\cite{hinton2015distilling} proposes to match the output probability distributions of two models by minimizing the Kullback-Leibler (KL) divergence loss. To improve distillation performance, existing methods have designed various forms of knowledge transfer. The method in FitNet~\cite{romero2014fitnets} proposes to let the student mimic the intermediate representations of the teacher network. AT~\cite{zagoruyko2016paying} tries to transfer the attention map of teacher to student. FSP~\cite{yim2017gift} proposes to generate the FSP matrix from the layer feature and use this matrix to guide the learning of student. The paraphraser and translator network is introduced in FT~\cite{kim2018paraphrasing} to aid the knowledge transfer process.
Many works~\cite{xu2019data,tung2019similarity,yun2020regularizing} have also explored the relationship between data samples in distillation. RKD~\cite{park2019relational} proposes a distillation framework that penalizes structural differences in relationships based on distance and angle. CCKD~\cite{peng2019correlation} introduces a kernel-based method for capturing complex inter-instance correlations.
In addition to the traditional classification task, knowledge distillation was also used to effectively obtain a light-weight network when training and designing networks for object detection~\cite{li2017mimicking,chen2017learning}, semantic segmentation~\cite{liu2019structured,Ye2019StudentBT}, and human pose estimation~\cite{zhang2019fast,li2021online}.

Unlike the two-stage offline knowledge distillation methods we described above, online counterparts reduces the training process to one-stage by eliminating the need for a large pre-trained teacher network. DML~\cite{zhang2018deep} proposes a novel student-teaching-student paradigm which conducts distillation collaboratively for peer student networks by learning from each other. DCM~\cite{yao2020knowledge} adds multiple auxiliary classifiers to the certain hidden layers of student networks in the two-way distillation method. ONE~\cite{zhu2018knowledge} trains a multi-branch student network while the teacher is established on-the-fly. OKDDip~\cite{chen2020online} introduces the concept of two-level distillation and uses the self-attention mechanism~\cite{vaswani2017attention} to build diverse peer networks. KDCL~\cite{guo2020online} removes the shared low-level structures and enables models of various capacities to learn collaboratively in the ensemble method. Li \textit{et al.}~\cite{li2020online} further improves the ensemble-based distillation method by enhancing the branch diversity. Walawalkar \textit{et al.}~\cite{walawalkar2020online} proposes to simultaneously train multiple student networks with different compression rates in one training procedure. 

\noindent\textbf{Self-Knowledge Distillation} Self-knowledge distillation can be approximately divided into two parts: data augmentation based approaches~\cite{xu2019data,yun2020regularizing} and auxiliary branch based approaches~\cite{zhang2019your,ji2021refine}. 
The data augmentation based method transfers the knowledge between the different augmented versions of the same training data without the need for additional teacher network. Xu \textit{et al.}~\cite{xu2019data} proposes to learn the consistent feature or posterior distributions between the different augmented versions of the same training samples. CS-KD~\cite{yun2020regularizing} proposes the class-wise regularization term that penalizes the predictions between different samples of the same label.

The auxiliary branch based self-distillation method introduces multiple convolutional branches on the backbone network to exploit its own knowledge. By simplifying the structure of the auxiliary branches, the self-distillation method achieves higher distillation efficiency than the online counterparts. DKS~\cite{sun2019deeply} first explores the possibility of utilizing the knowledge learned by the auxiliary branch to regularize the training of the backbone network. The backbone network is treated as the teacher and student at the same time. In BYOT~\cite{zhang2019your}, the backbone network is divided into several blocks according to its structures and depth. Additional bottleneck and fully connected layers are set after each block, which constitutes the auxiliary classifier. Each auxiliary classifier in the network is treated as a student. During training, the backbone network plays the role of a teacher, providing the high-accuracy soft labels to train the multiple student sub-networks. BYOT typically follows the classical teacher-teaching-student paradigm and all well-trained students can be utilized independently to meet the needs of different compression rates. But it neglects that the teacher network is also being optimized during training. The trained backbone teacher network shows superior performance to the baseline method. This gives us the inspiration that the teacher can actually learn from students and get meaningful improvements.
Inspired by FPN~\cite{lin2017feature}, FRSKD~\cite{ji2021refine} constructs an auxiliary self-teacher network that distills the refined feature maps to the original classifier. The soft label generated by the self-teacher is also utilized for the distillation. A major drawback of the above work is that they neglect the advantages of the diverse supervision signals provided by multiple auxiliary classifiers.

\begin{figure*}[t]
	\centering
	\includegraphics[width=1\linewidth]{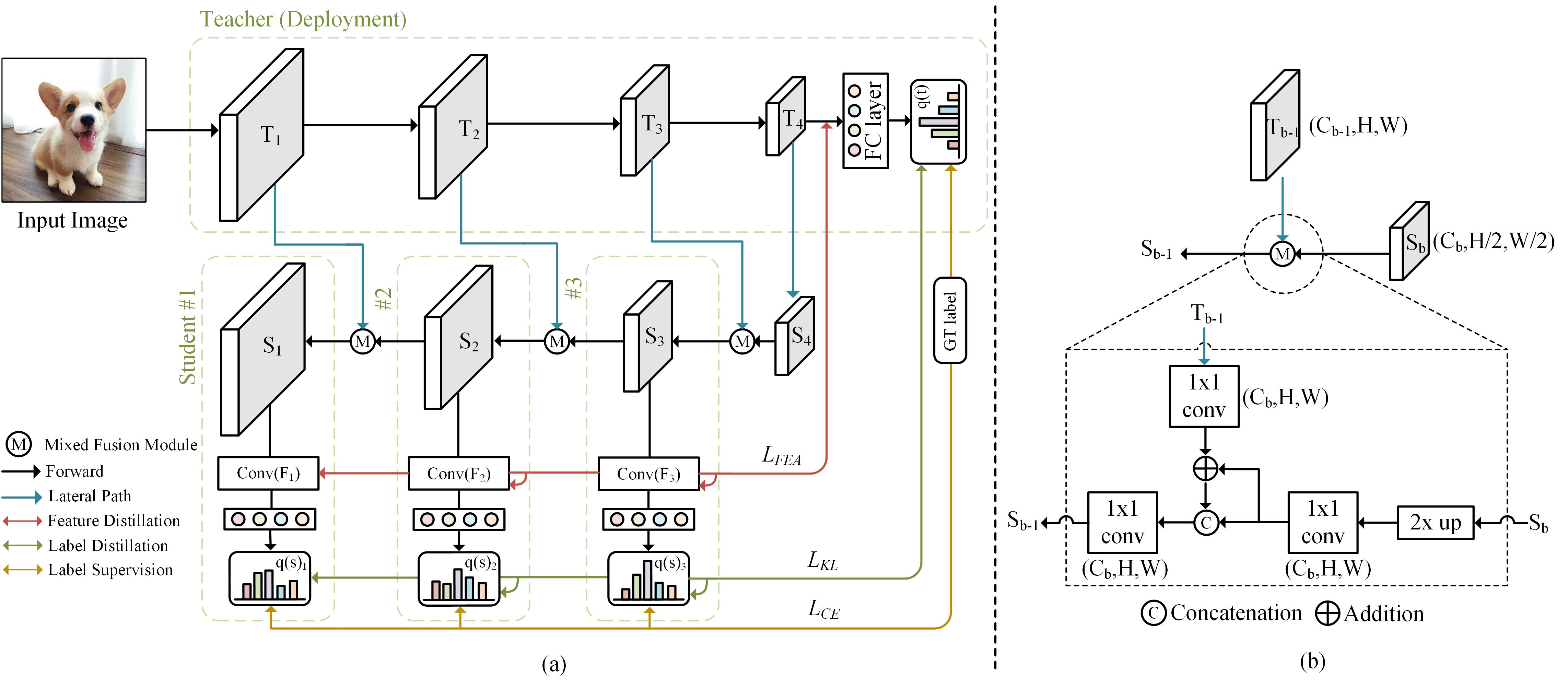}
	\caption{
		An overview of the our proposed student-helping-teacher method, Teacher Evolution via Self-Knowledge Distillation (TESKD). We divide the target backbone teacher into four blocks and construct three hierarchical student sub-networks $\#1$, $\#2$ and $\#3$ in a FPN-like way by sharing various stages of the teacher backbone features. When the teacher provides high-quality soft labels to guide the hierarchical student, it also make improvements based on students' diverse feedback via the shared intermediate representations.
	}
	\label{fig:architecture}
\end{figure*}

\section{Teacher Evolution via Self-Knowledge Distillation}
In this section, we first briefly review the basic concept of the classical knowledge distillation that follows the \textit{teacher-teaching-student} paradigm. Then we propose our novel \textit{student-helping-teacher} framework and introduce the mixed fusion module.

\subsection{Background and Notations}

Knowledge distillation~\cite{hinton2015distilling}, as one of the main network compression techniques~\cite{wu2016quantized,molchanov2016pruning}, has been widely used in many tasks~\cite{li2017mimicking,liu2019structured,zhang2019fast}. The traditional two-stage distillation process usually starts with a pre-trained cumbersome teacher network. Then a compact student network will be trained under the supervision of the teacher network in the form of soft predictions or intermediate representations~\cite{romero2014fitnets,yim2017gift}. After the distillation, the student can master the expertise of the teacher and thus is used for efficient deployment.
Such a learning process can be typically viewed as a \textit{teacher-teaching-student} paradigm. 
% After the two-stage distillation, the well-trained student is used for final deployment. 
Given the labeled classification dataset $D=\{(\textbf{x}_{i},\textbf{y}_{i})\}_{i=1}^{n}$, the Kullback-Leibler (KL) divergence loss is used to minimize the discrepancy between the soften output probabilities of the student network and teacher network:
\begin{equation}
L_{KL} = \sum_{i=1}^{n}T^{2} KL(q(s)_{i}, q(t)_{i}),
\end{equation}
where $T$ is the temperature parameter to scale the smoothness of distribution, $q(s)$ and $q(t)$ denotes the soften probability produced by the student and the teacher, respectively. The predicted probabilities are calculated with a softmax layer built on logits $t_{i}$, i.e., $q_{i}=softmax(t_{i}/T)$. A larger temperature $T$ will make the probability distribution softer.

To train a multi-class classification network, we also minimize the traditional Cross-Entropy (CE) loss between the predicted probabilities $q(s)_{i}$ and the ground-truth one-hot label $y_{i}$ of each training sample:

\begin{equation}
L_{CE} = \sum_{i=1}^{n}CE(q(s)_{i}, y_{i}),
\end{equation}

With both hard labels and soft labels, the final loss function of the conventional knowledge distillation is written with the balancing parameter $\lambda$ as follows:

\begin{equation}
L_{total}=L_{CE}+\lambda L_{KL},
\end{equation}

\subsection{Our Method}
As opposed to traditional distillation approaches that follow the \textit{teacher-teaching-student} paradigm,
% Different from the traditional \textit{teacher-teaching-student} paradigm,
in this work, we explore a novel and reverse aspect that studies the \textit{student-helping-teacher} scheme, aiming at improving the teacher network through the effective feedback from multiple students. An overview of our proposed distillation framework is illustrated in Fig.~\ref{fig:architecture}. The whole framework mainly consists of two components: 

\noindent\textbf{(1)} The backbone teacher network $T$ with $B$ stages for deployment, which meets the requirement of the target complexity.
%The target backbone teacher is divided into $B$ blocks according to its structures and depth. 
Note that it is different from the most existing practices that usually train student network for deployment.

\noindent\textbf{(2)} $B-1$ auxiliary hierarchical student sub-networks $S$ constructed with a top-down architecture, each of which shares the feature maps from the corresponding teacher backbone stages.

Specifically, given an input RGB image, we can obtain the backbone teacher feature set $T=\{T_{1}, T_{2}, ..., T_{B}\}$ and the output probability distribution $q(t)$ after the feed forward computation. The student sub-network takes the last teacher feature $T_{B}$ as initial input and obtain $S_{B}$. For every $1\le b\le B-1$, it generates higher resolution features $S_{b}$ iteratively by upsampling spatially coarser, but semantically stronger feature maps $S_{B}$. Then these features $S_{b}$ are fused with semantically weaker, but spatially finer, features $T_{b-1}$ via the proposed Mixed Fusion Module (MFM). The details of MFM are elaborated in the following section. In this way, we can obtain the hierarchical student feature set $S=\{S_{1}, S_{2}, ..., S_{B}\}$. Additional convolutional blocks and fully connected layers are set after each student block, each of which serves as an independent student classification model to generate the soft probability $q(s)_{b}$ for distillation.

The backbone teacher network has two learning tasks during distillation. It not only learns to generate high-accuracy labels for prediction and distillation, but also tries to provide robust intermediate features 
%to multiple hierarchical students.
to guide the learning of multiple hierarchical students.
Through such optimization, the backbone network can learn more generalized features and significantly outperforms the baseline network.
In test, multiple auxiliary student sub-networks can be simply removed while keeping the well-trained teacher network for deployment.

\subsection{The Mixed Fusion Module}
Inspired by MLN~\cite{Wang2018MixedLN}, we propose the Mixed Fusion Module (MFM) which consists of both addition and concatenation operators, leading to the multiple hierarchical student sub-networks with a top-down architecture. The architecture of our proposed MFM is depicted in Fig.~\ref{fig:architecture}(b).
% The Mixed Fusion Module (MFM) utilizes the insight of MLN~\cite{Wang2018MixedLN}, the details are shown in Fig.~\ref{fig:architecture}(b).

Previous works~\cite{lin2017feature,ronneberger2015u} use addition and concatenation operation independently to fuse the features from the encoder (backbone) and decoder network. But simply adding or concatenating two features may impede the information flow or bring redundancy to the network, resulting in performance degradation. In this work, we mix these two operations to effectively bridge the information flow between the backbone teacher and multiple hierarchical students, in order to combine both the advantages of them and avoid the possible limitations.

% \rewrite{check here}. % 图上用 S_i 为什么这里用 S_b ? 保持一致吧，论文中的所有标记要与图中的标记保持一致；可以把一些段落中定义的函数 例如 f_s(\cdot) 写到图中对应位置便于reviewer查看
Specifically, with a spatially coarser feature $S_{b}$, we upsample the spatial resolution by a factor of 2. The following $1\times1$ convolution operation is used to align the channel dimension between two input features. Then the upsampled features are fused with the corresponding student features $S_{b-1}$ through the addition and concatenation operations, which can be formulated as Eqn.~(4):

\begin{equation}
\begin{aligned}
S_{b}= Conv(f_{t}(T_{b})+ f_{s}(S_{b+1})) \\ \parallel  ~f_{s}(S_{b+1}), b \in [1, B-1],
\end{aligned}
\end{equation}
where the symbol ``$+$" and ``$\parallel$" denotes addition and concatenation operation, respectively. $f_{s}(\cdot)$ is the function of upsampling and the $1\times1$ convolution block. $f_{t}(\cdot)$ is the function of the $1\times1$ convolution block, which is used to align the number of channels between two features.
$Conv$ is a $1\times1$ convolution block, which fuse the features after the concatenation operation and halves the number of feature channels to $C_{b}$, as shown in Fig.~\ref{fig:architecture}(b). This fusion process is iterated until the latest feature map $S_{1}$ is generated. 
% Since the student features are constructed based on the backbone teacher features, 
We set the channel dimension of all student features to $C_{B}$, which is the same channel dimension as the last teacher feature $T_{B}$.

% The branch-based self-distillation method usually relies on the accuracy of the soft labels generated by the branch classifier. 
Through our proposed MFM, even the shallowest student sub-network can still obtain sufficient spatial and semantic information at the same time, resulting in better representation ability.
Multiple stronger student networks can provide more diverse and sufficient feedback signals, from which the backbone teacher can learn and be significantly improved.

%Since multiple auxiliary students shares various stages of the teacher features, higher accuracy of auxiliary student sub-networks empirically indicates better backbone teacher representation ability. 

% Its accuracy will significantly affect the efficacy of knowledge transfer since the auxiliary labels serves as teachers by regulating students' learning process. Through the iterative fusion of multi-scale features via mixed fusion module, the features $T$ fed into the auxiliary branch have richer spatial and semantic information, which results in better branch accuracy and improves the network performance.

\subsection{Feature Distillation}

After the feed forward computation through the FPN-like multi-branch distillation framework, we can obtain the proportionally sized student's feature set $S=\{S_{1}, S_{2}, ..., S_{B}\}$, teacher's feature set $T=\{T_{1}, T_{2}, ..., T_{B}\}$ and branches' intermediate feature set $F=\{F_{1}, F_{2}, ..., F_{B-1}\}$. Specifically, we aim to make the backbone teacher classifier guide the learning process of all auxiliary student classifiers. The guided layer is selected as the last layer of the convolutional block in the auxiliary branch.
%Feature maps of a teacher model are valuable for learning a better student model as depicted in ~\cite{romero2014fitnets}. 
We minimize the L2 loss between the intermediate feature maps in the main teacher classifier and other auxiliary student classifiers, which can be written as:
\begin{equation}
L_{FEA}= \sum_{b=1}^{B-1} ||F_{b}-T_{B}||_{2}^{2},
\end{equation}

Note that the convolution blocks in student sub-network are made of regular convolution layers, batch normalization~\cite{ioffe2015batch} and ReLU activation function.  For each student, instead of the bottleneck design in BYOT~\cite{zhang2019your}, we construct a simple single conv-bn-relu block to first reduce the feature map size (e.g., to $4\times4$ for CIFAR and $7\times7$ for ImageNet). Then a global average pooling layer is applied for feature distillation.
% The convolutional kernel size is adjusted based on the size of the input features in order to effectively capture the global context and improve generalization ability. 
% \rewrite{check here:} % 这里不用讲清楚具体怎么adjusted的么，大家都默认知道？

\textbf{Overall} To get a better understanding of our proposed TESKD, the full training procedure is summarized in Algorithm~\ref{algo:demo}. The overall loss of our proposed self-distillation framework is given as:

\begin{equation}
\begin{aligned}
L_{total}= & \alpha_{1} \sum_{b=1}^{B} L_{CE}(q_{i},y) + \\ & \alpha_{2} \sum_{b=1}^{B-1}L_{KL}(q(t)_{b},q(s)) + \beta L_{FEA}.
\end{aligned}
\end{equation}
where $\alpha_{1}$, $\alpha_{2}$ and $\beta$ are the hyperparameters to control the impact of each loss term which also satisfy $\alpha_{1}+\alpha_{2}=1$. $q$ denotes all the predicted results generated by the whole network, i.e. $\{q(t),q(s)_{1},q(s)_{2},q(s)_{3}\}$. The first loss term is the total cross-entropy loss to the ground truth labels of both all auxiliary students and the backbone teacher network.

\begin{algorithm}[t]
	\caption{Student Helping Teacher: Teacher Evolution via Self-Knowledge Distillation.}
	\hspace*{0.02in} {\bf Input:}
	Labelled Training dataset $D=\{(\textbf{x}_{i},\textbf{y}_{i})\}_{i=1}^{n}$; Training Epoch Number $\epsilon$; A target teacher model $\theta^{t}$; Three hierarchical student models $\theta^{s}$; \\
	\hspace*{0.02in} {\bf Output:}
	A well-trained target teacher model; \\
	\hspace*{0.02in} {\bf Initialize:}
	Epoch e=1; Randomly initialize $\theta^{s}$ and $\theta^{t}$;
	
	\begin{algorithmic}[1]
		\While {e $\leq$ $\epsilon$}
		\State Forward propagation and obtain the intermediate features in teacher and multiple students. (Eqn.~(4));
		\State Compute the predictions of all auxiliary students $q(s)$ and teacher $q(t)$.;
		\State Compute the Cross-Entropy loss $L_{CE}$. (Eqn.~(2));
		\State Distill the knowledge from the teacher model to multiple hierarchical student models. (Eqn.~(1));
		\State Align the feature representations between teacher and multiple students. (Eqn.~(5));
		\State Obtain the final loss function. (Eqn.~(6));
		\State Update the model parameters $\theta^{s}$ and $\theta^{t}$.;
		\State e=e+1
		\EndWhile
		\State \textbf{end while}
	\end{algorithmic}
	\label{algo:demo}
\end{algorithm}

\begin{table*}[t]
	\begin{center}
		\begin{tabular}{c|ccccccc}
			\hline\noalign{\smallskip}
			Models & VGG-16  & VGG-19	& ResNet-18 & ResNet-34 & SENet-34 & ResNeXt-18 & ResNeXt-34 \\
			\noalign{\smallskip}
			\hline
			\noalign{\smallskip}
			Baseline& 72.70 $\pm$ 0.24 	& 72.81 $\pm$ 0.25	& 74.40 $\pm$ 0.08	& 74.51 $\pm$ 0.16	& 74.54 $\pm$ 0.14	& 
			75.38 $\pm$ 0.17 & 76.08 $\pm$ 0.06	\\
			ONE	    & 73.24 $\pm$ 0.10	& 72.13 $\pm$ 0.11	& 77.01 $\pm$ 0.28	& 77.24 $\pm$ 0.23	& 76.19 $\pm$ 0.15	& 
			77.42 $\pm$ 0.09 & 78.25 $\pm$ 0.14	\\
			CS-KD   & 73.43 $\pm$ 0.14	& 73.13 $\pm$ 0.10	& 77.26 $\pm$ 0.14	& 76.82 $\pm$ 0.29	& 76.85 $\pm$ 0.18	&
			78.32 $\pm$ 0.26 & 77.35 $\pm$ 0.11	\\
			BYOT 	& 73.70 $\pm$ 0.25  & 74.18 $\pm$ 0.19	& 77.22 $\pm$ 0.20 	& 77.93 $\pm$ 0.06  & 77.58 $\pm$ 0.14	& 
			74.28 $\pm$ 0.21 & 75.23 $\pm$ 0.27	\\
			FRSKD   & 69.24 $\pm$ 0.05	& 70.38 $\pm$ 0.11	& 77.51 $\pm$ 0.15	& 77.58 $\pm$ 0.09 	& 77.46 $\pm$ 0.18	& 
			78.13 $\pm$ 0.03 & 77.11 $\pm$ 0.11	\\
			\noalign{\smallskip}
			\hline
			\noalign{\smallskip}
			Ours w/o F &	74.73 $\pm$ 0.20		  &	74.44 $\pm$ 0.14		  & 78.61 $\pm$ 0.09		  &	78.90 $\pm$ 0.15
			& 	78.69 $\pm$ 0.06	  &	79.57 $\pm$ 0.14		  &	79.49 $\pm$ 0.23\\
			Ours  				  & \textbf{74.90 $\pm$ 0.17} &	\textbf{75.01 $\pm$ 0.22} & \textbf{79.14 $\pm$ 0.11} & \textbf{79.60 $\pm$ 0.16} & \textbf{78.97 $\pm$ 0.06} & \textbf{79.65 $\pm$ 0.12} & \textbf{79.77 $\pm$ 0.08} \\
			\noalign{\smallskip}
			\hline
		\end{tabular}
		\caption{
			Accuracy (\%) comparison of various distillation approaches on CIFAR-100 dataset. ``w/o F" denotes without feature distillation. The best performing model is indicated as boldface. 
		}
		\label{table:cifar100}
	\end{center}
\end{table*}

\section{Experiments}

We evaluate our proposed TESKD framework on four popular neural networks (VGG~\cite{simonyan2014very}, ResNet, SENet~\cite{hu2018squeeze}, ResNeXt~\cite{xie2017aggregated}) and two benchmark datasets~(CIFAR-100~\cite{krizhevsky2009learning}, ImageNet-2012~\cite{deng2009imagenet}). We compare our method with closely related self-distillation and online distillation works (ONE~\cite{zhu2018knowledge}, CS-KD~\cite{yun2020regularizing}, FRSKD~\cite{ji2021refine} and BYOT). Detailed ablation studies on the network components are also conducted to demonstrate its effectiveness. All evaluations are made in comparison to state-of-the-art approaches based on standard experimental settings and reported in means and standard deviations over 3 runs.

\textbf{Dataset.}
The CIFAR-100 dataset consists of colored natural images with $32\times32$ pixels. The training and testing sets contain 50K and 10K images, respectively. Same as previous works~\cite{zhang2019your,ji2021refine}, the network structures are modified to fit the tiny images in CIFAR-100.

The ImageNet-2012 classification dataset is more challenging than CIFAR. It contains 1.2M images for training, 50K for validation, from 1K classes. The resolution of input images after pre-processing is $224\times224$.

\textbf{Implementation details.}
All the methods are implemented by PyTorch~\cite{paszke2019pytorch}. For CIFAR-100, we follow the standard data augmentation scheme for all training images as in~\cite{zhang2019your,ji2021refine}, i.e. random cropping and horizontal flipping. We use the stochastic gradient descents~(SGD) as the optimizer with momentum 0.9 and weight decay 5e-4 during training. The learning rate starts from 0.1 and is divided by 10 at 100 and 150 epochs, for a total of 200 epochs. For ImageNet, we set the intital learning rate to 0.1 and divide the learning rate by 10 at 30 and 60 epochs, for a total of 90 epochs. Weight deacy is set to 1e-4. Same data augmentation scheme is adopted as in \cite{ji2021refine}. The mini-batch size is 128 and 256 for CIFAR and ImageNet, respectively. We set $\alpha_{1}$ to 0.2 and 0.8 in Eqn.~(6) for CIFAR-100 and ImageNet, respectively. $\beta$ is usually set to 1e-7. We set $B$ to $4$ for all methods.

\subsection{Experiments on CIFAR-100}

Table~\ref{table:cifar100} shows the top-1 classification accuracy on CIFAR-100 based on seven varying capacity state-of-the-art neural networks. Note that we set the branch number in ONE to 3 and the reported results are the averaged accuracy of all branches. For BYOT, we report the results of the main network (i.e. backbone network). From this table, we can observe that all different networks benefit significantly from our proposed TESKD, particularly for small models achieving larger performance improvements. For example, the ResNet-18$/$ResNeXt-18 network trained with our method shows 78.61\%$/$79.57\% accuracy, outperforming the baseline by 4.74\%$/$4.27\% margin. Our method obtains an average of 3.8\% improvement on different baseline networks and surpasses several closely related state-of-the-art self-distillation methods with obvious margins. Two distillation terms exist in our method: label distillation and feature distillation. Since some comparison methods do not use the feature distillation technique, we remove this term for fair comparison. 'Ours w/o F' denotes the results we do not perform the feature distillation during training. When we simply remove the feature distillation term, our results still show superior performance even compared with other state-of-the-art self-distillation methods. This indicates that the supervision signals provided by the soft labels are the key to the self-distillation method.

\begin{figure}
	\centering
	\includegraphics[width=1\linewidth]{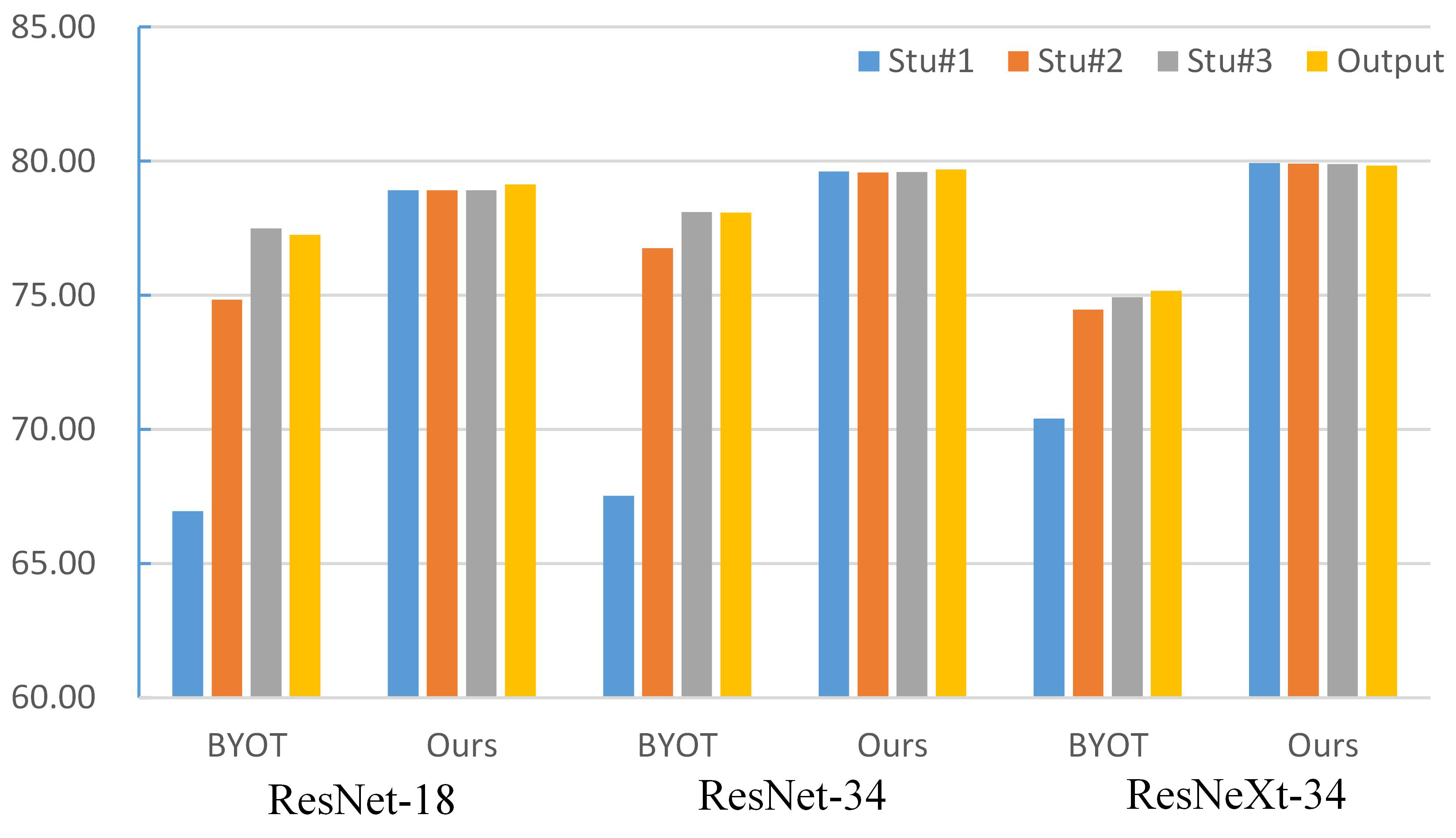}
	\caption{
        Top-1 accuracy (\%) comparison of student sub-networks on CIFAR-100 dataset. ``Stu" is the abbreviation of ``student".
	}
	\label{fig:data_visualization}
\end{figure}

BYOT also introduces multiple student branch networks that share the underlying backbone network. After the distillation, the teacher's performance has also been improved, as we shown in Table~\ref{table:cifar100}. But the performance improvement of the teacher network in BYOT is not high. This is because it ignores the quality of the student network. The student network only shares a limited number of underlying convolutional blocks, resulting in insufficient representation ability. The teacher network cannot obtain sufficient and effective feedback from multiple students during optimization. We compare the classification accuracy of multiple student networks between our proposed TESKD and BYOT, as shown in Fig.~\ref{fig:data_visualization}. We can see that for three different backbone networks, there is an obvious performance gap between three student networks in BYOT. The performance of the student $\#1$ is the worst, since it only shares the shallowest layer of the backbone network. But in our TESKD, through our proposed top-down pathway, rich semantic information can be transmitted to the bottom, so that even the shallowest student network can still get a better classification accuracy. All the student networks of our method show superior performance than BYOT. This shows that high-quality students can help teachers achieve more significant improvement.

\subsection{Experiments on ImageNet}

Table~\ref{table:imagenet} shows the results on a large-scale image classification dataset ImageNet-2012~\cite{deng2009imagenet} based on ResNet-18. We compare our method with two closely related state-of-the-art self-knowledge distillation approaches on ImageNet. The ResNet-18 network trained with our proposed method shows 71.14\% accuracy, outperforming the baseline by 1.43\% margin. This demonstrates that our method can still be applied to large-scale dataset effectively.

% 分支的三个acc对比，后面会画成柱状图。
% \begin{table}
% 	\begin{center}
% 		\resizebox{1\linewidth}{!}
% 		{
% 			\begin{tabular}{c|c|cccc}
% 				\hline\noalign{\smallskip}
% 				Network  & Method       	& Stu \#1 & Stu \#2 & Stu \#3 & Output\\
% 				\noalign{\smallskip}\hline\noalign{\smallskip}
% 				\multirow{2}*{ResNet-18} 	& BYOT	& 66.96 & 74.83 & 77.49 & 77.26 \\
% 				~						 	& Our   & 78.91 & 78.91 & 78.92 & \textbf{79.13} \\
% 				\noalign{\smallskip}\hline\noalign{\smallskip}
% 				\multirow{2}*{ResNet-34} 	& BYOT	& 67.53	& 76.76	& 78.10	& 78.09 \\
% 				~						 	& Our   & 79.61 & 79.58 & 79.60 & \textbf{79.70} \\
% 				\noalign{\smallskip}\hline\noalign{\smallskip}
% 				\multirow{2}*{ResNeXt-34} 	& BYOT	& 70.40	& 74.47	& 74.93	& 75.16 \\
% 				~						 	& Our   & 79.94	& 79.91	& 79.89	& \textbf{79.84} \\
% 				\hline
% 			\end{tabular}
% 		}
% 		\caption{Top-1 accuracy (\%) comparison of student sub-networks on CIFAR-100 dataset.  Stu is the abbreviation of student.}
% 		\label{table:branch_compare}
% 	\end{center}
% \end{table}

\begin{table}
	\begin{center}
		\resizebox{0.7\linewidth}{!}
		{
			\begin{tabular}{cccc}
				\hline\noalign{\smallskip}
				Baseline & BYOT  & FRSKD & Ours  \\
				\noalign{\smallskip}\hline\noalign{\smallskip}
				69.71    & 69.84 & 70.17 &	71.14	\\
				\noalign{\smallskip}\hline
			\end{tabular}
		}
		\caption{Top-1 classification accuracy (\%) comparison with other SOTA self-distillation methods for ResNet-18 on ImageNet-2012 dataset.}
		\label{table:imagenet}
	\end{center}
\end{table}

\subsection{Compared with Traditional Distillation}

\begin{table*}[t]
	\begin{center}
		\resizebox{1\linewidth}{!}
		{
			\begin{tabular}{c|c|ccccccc}
				\hline\noalign{\smallskip}
				Teacher		& Student 	& Baseline 			& KD 	            & FitNet 	        & AT	            & SP	                & VID		 & Ours 		\\
				\noalign{\smallskip}\hline\noalign{\smallskip}
				ResNet-101	& ResNet-18	& 74.40 $\pm$ 0.08	& 77.12 $\pm$ 0.24  & 77.38 $\pm$ 0.18	& 77.41 $\pm$ 0.25	& 77.84 $\pm$ 0.14     & 77.34 $\pm$ 0.18  & 79.14 $\pm$ 0.11	\\
				ResNet-101	& ResNet-34	& 74.51 $\pm$ 0.16	& 77.70 $\pm$ 0.21	& 77.97 $\pm$ 0.06	& 77.43 $\pm$ 0.16	& 77.50 $\pm$ 0.05	    & 77.73 $\pm$ 0.14  & 79.60 $\pm$ 0.16	\\
				\noalign{\smallskip}\hline\noalign{\smallskip}
				ResNeXt-152	& ResNeXt-18 & 75.38 $\pm$ 0.17	& 78.81 $\pm$ 0.18 	& 79.08 $\pm$ 0.08  & 78.85 $\pm$ 0.13	& 78.84 $\pm$ 0.16		 & 78.72 $\pm$ 0.13	 & 79.65 $\pm$ 0.12	\\
				ResNeXt-152 & ResNeXt-34 & 76.08 $\pm$ 0.06	& 78.65 $\pm$ 0.06  & 78.77 $\pm$ 0.16	& 78.72 $\pm$ 0.19	& 78.68 $\pm$ 0.13		 & 78.58 $\pm$ 0.08	 & 79.77 $\pm$ 0.08	\\
				\noalign{\smallskip}\hline
			\end{tabular}
		}
		\caption{Top-1 accuracy comparison with traditional distillation method on CIFAR-100 dataset.}
		\label{table:traditional_kd_comparison}
	\end{center}
\end{table*}

% \begin{table*}
% 	\begin{center}
% 		%		\resizebox{1\linewidth}{!}
% 		{
% 			\begin{tabular}{c|ccccc}
% 				\hline\noalign{\smallskip}
% 				Method          & w/o MFM (No Connect)  & w/o MFM (Add) & w/o MFM (Concat)  & w/o KD 			& TESKD \\
% 				\noalign{\smallskip}\hline\noalign{\smallskip}
% 				Top-1 Acc 	& 78.40 $\pm$ 0.13 	  & 78.77 $\pm$ 0.10 & 78.63 $\pm$ 0.09 	& 77.25 $\pm$ 0.22	& \textbf{79.14 $\pm$ 0.11}	\\
% 				\noalign{\smallskip}\hline
% 			\end{tabular}
% 		}
% 		\caption{Ablation Study: Accuracy comparison (\%) of different connection operations. (ResNet-18 on CIFAR-100 dataset)}
% 		\label{table:mixed_link}
% 	\end{center}
% \end{table*}

\begin{table}
	\begin{center}
		\resizebox{0.85\linewidth}{!}
		{
			\begin{tabular}{lc}
				\hline\noalign{\smallskip}
				Method & Top-1 Acc \\
				\noalign{\smallskip}\hline\noalign{\smallskip}
				w/o MFM-No Connection             & 78.40 $\pm$ 0.13      \\ 
				w/o MFM-Concat (i.e., FPN Style)  & 78.63 $\pm$ 0.09      \\
				w/o MFM-Add (i.e., UNet Style)    & 78.77 $\pm$ 0.10      \\
				w/o Knowledge Distillation        & 77.25 $\pm$ 0.22      \\
				\noalign{\smallskip}\hline\noalign{\smallskip}
				TESKD                            & 79.14 $\pm$ 0.11      \\
				\noalign{\smallskip}\hline
			\end{tabular}
		}
		\caption{Ablation Study: Impact of different connection operations in TESKD. (ResNet-18 on CIFAR-100 dataset)}
		\label{table:mixed_link}
	\end{center}
\end{table}

Traditional knowledge distillation methods usually follow the teacher-teaching-student paradigm. Table~\ref{table:traditional_kd_comparison} compares the classification accuracy of our proposed method with five traditional distillation methods on CIFAR-100, including vanilla KD, FitNet~\cite{romero2014fitnets}, AT~\cite{zagoruyko2016paying}, SP~\cite{tung2019similarity} and VID~\cite{ahn2019variational}. The results of the vanilla KD are also included for comparison. Each column includes the results of corresponding student models which are generated by the supervision of the same teacher. According to Table~\ref{table:traditional_kd_comparison}, it is shown that our proposed method consistently achieves higher accuracy than the state-of-the-art traditional distillation approaches with an extra teacher network.

Traditional knowledge distillation methods need to train a cumbersome teacher network at first, then distill the knowledge to the light-weight student network. Such a two-stage learning procedure is complex and time-consuming. Instead, we train the student and teacher network simultaneously in a one-stage manner, eliminating the need for the pre-trained teacher. We have achieved higher training efficiency and accuracy than traditional methods.

% 在这里的ablation study最后可以加一个different proportions of CE and KD loss weight. 因为全文里的权重CE+KD=1，在cifar上面可以列出来 0.2/0.8, 0.4/0.6, 0.6/0.4, 0.8/0.2 这样的四个实验，参考Revisiting Knowledge Distillation: An Inheritance and Exploration Framework的Table 5的方式。不同的dataset和backbone都可以试一试。
\subsection{Ablation Study}

% \subsubsection{Study on different connection operations}

The Mixed Fusion Module (MFM) consists of both addition and concatenation operators, which sufficiently bridge the information flow between the teacher and multiple students. To further evaluate the effectiveness of each individual component in our proposed TESKD, especially for the Mixed Fusion Module, we perform various ablation studies on CIFAR-100 based on ResNet-18, as shown in Table~\ref{table:mixed_link}. Specifically, we compare the performance of TESKD with the following four ways of ablations: 

(1) \textbf{w/o MFM-No Connection.} We directly upsample the features $T_{4}$ from the last block of the teacher network and do not merge with other lateral features. This leads to a lower accuracy by 0.74\% (79.14\%-78.40\%).

(2) \textbf{w/o MFM-Concat (i.e., FPN Style).} Similar to FPN, we merge the feature maps of the same spatial size from the backbone teacher network and the upsampled student features without concatenation operation. Simply adding two features may impede the information flow and reduce the performance by 0.37\%.

(3) \textbf{w/o MFM-Add (i.e., UNet Style).} Inspired by UNet, the concatenation operation is applied to merge the features from two ways. Another $1\times1$ convolutional layer is added to halve the number of feature channels after the concatenation operation. However, the concatenation may bring redundancy when it concats too many raw features from the backbone teacher network, resulting in a 0.51\% performance degradation.

(4)\textbf{w/o Knowledge Distillation.} We directly train the entire multi-branch network without any distillation terms. This increase the error rate by 1.89\%, which confirms the effectiveness of our proposed one-stage self-knowledge distillation framework.

\begin{table}
	\begin{center}
		\resizebox{0.7\linewidth}{!}{
    		\begin{tabular}{cccc}
    			\hline\noalign{\smallskip}
    			Stu $\#1$ & Stu $\#2$ & Stu $\#3$  & Top-1 Acc \\
    			\noalign{\smallskip}
    			\hline
    			\noalign{\smallskip}
    			& 		 	 & \checkmark & 78.37 $\pm$ 0.27  \\
    			& \checkmark & \checkmark &	78.64 $\pm$ 0.13	\\
    			\checkmark  & \checkmark & \checkmark & 79.14 $\pm$ 0.11 \\
    			\noalign{\smallskip}
    			\hline
    		\end{tabular}
    	}
    	\caption{
    		Ablation Study: Impact of the student sub-networks with ResNet-18 on CIFAR-100 dataset.
    	}
		\label{table:student_numbers}
	\end{center}
\end{table}

\subsection{Impact of the student sub-networks}

We evaluate the impact of the student sub-networks on the performance of our branch-based self-distillation approach. As we shown in Fig.~\ref{fig:architecture}, we have three sub-networks in our proposed framework. When we reduce the number of student sub-networks, the diversity of feedback that students provide to the teacher will also decrease. This will affect the teacher's performance. The results are summarized in Table~\ref{table:student_numbers}. Multiple student sub-networks are removed one-by-one to measure their effect. Noted that if we remove all students, our method will not be able to perform the distillation operation, so we keep the last student $\#3$. From Table~\ref{table:student_numbers}, we can observe that when we remove the auxiliary students one-by-one, the performance of the teacher model used for deployment gradually decreases. This verifies our idea that feedback from multiple students can indeed affect teacher's learning.

\section{Conclusion}

Different from the existing teacher-teaching-student and student-teaching-student paradigm, in this paper, we propose a novel student-helping-teacher formula, Teacher Evolution via Self-Knowledge Distillation (TESKD) where the target  teacher is learned with the help of multiple hierarchical students by sharing structural backbone. The well-trained teacher is used for final deployment.
Extensive experiments have validated the effectiveness of our proposed TESKD on two popular benchmark datasets.

\bibliography{aaai22}

\end{document}